\documentclass[10pt,twocolumn,letterpaper]{article}

\usepackage{paper}              

\definecolor{myblue}{rgb}{0.21,0.49,0.74}
\usepackage[pagebackref,breaklinks,colorlinks,allcolors=myblue]{hyperref}
\usepackage{algorithm}
\usepackage{algorithmic}

\usepackage{multirow}
\def\method{ProgRoCC}

\title{\method: A Progressive Approach to Rough Crowd Counting}

\author{Shengqin Jiang\textsuperscript{1} \quad Linfei Li\textsuperscript{1} \quad Haokui Zhang \textsuperscript{2} \quad Qingshan Liu \textsuperscript{3} \quad Amin Beheshti \textsuperscript{4}\\
	  \quad Jian Yang \textsuperscript{4} \quad Anton van den Hengel \textsuperscript{5} \quad Quan Z. Sheng \textsuperscript{4} \quad Yuankai Qi \textsuperscript{4} \\
	  \\
\textsuperscript{1}School of Computer Science, Nanjing University of Information Science and Technology \\
\textsuperscript{2}School of Cybersecurity, Northwestern Polytechnical University \\
\textsuperscript{3}School of Computer Science, Nanjing University of Posts and Telecommunications \\
\textsuperscript{4} School of Computing, Macquarie University \\
 \textsuperscript{5}Australian Institute for Machine Learning, The University of Adelaide \\}

\begin{document}
\maketitle
\begin{abstract}
As the number of individuals in a crowd grows, enumeration-based techniques become increasingly infeasible and their estimates increasingly unreliable.
We propose instead an estimation-based version of the problem: we label Rough Crowd Counting that delivers better accuracy on the basis of training data that is easier to acquire. 
Rough crowd counting requires only rough annotations of the number of targets in an image, instead of the more traditional, and far more expensive, per-target annotations.
We propose an approach to the rough crowd counting problem based on CLIP, termed \method. Specifically, we introduce a progressive estimation learning strategy that determines the object count through a coarse-to-fine approach. 
This approach delivers answers quickly, outperforms the state-of-the-art in semi- and weakly-supervised crowd counting.
In addition, we design a vision-language matching adapter that optimizes key-value pairs by mining effective matches of two modalities to refine the visual features, thereby improving the final performance. 
Extensive experimental results on three widely adopted crowd counting datasets demonstrate the effectiveness of our method.

\end{abstract}    
\section{Introduction}
Estimation is a field of its own, somewhat related to quantity surveying. The task of an estimator is to provide as accurate an estimate of a quantity as possible in the case where direct measurement is infeasible.  
Direct crowd counting grows increasingly infeasible as the number of individuals rises.  People are increasingly occluded by other people, and other objects, either partially or in full. As crowd numbers increase, people get too small to be able to identify individually, leading to a bias towards underestimation in enumeration-based methods.  Facing similar challenges, professional estimators often use approaches like progressive-elaboration that build an answer in stages.

\begin{figure}[!t]
\centering
\includegraphics[width=0.99\columnwidth]{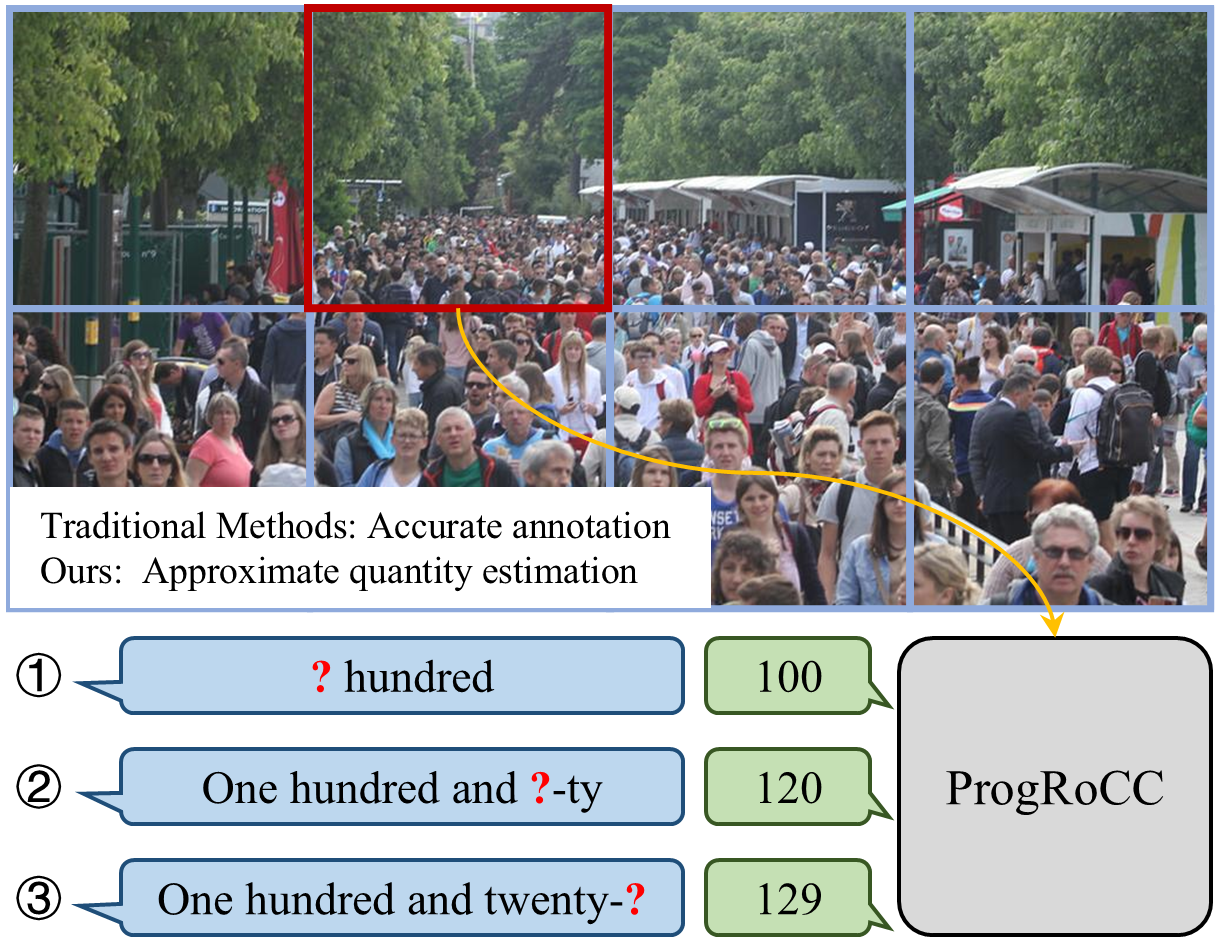}
\caption{Our solution to rough crowd counting. Compared to traditional counting methods that require accurate annotations, our method achieves robust crowd counting  using simpler approximate estimates of object numbers as labels. It captures the number of people through a progressive estimation learning strategy. }
\label{fig1}
\vspace{-4mm}
\end{figure}

With the continuous development of deep learning techniques, the accuracy and efficiency of computer vision methods for crowd counting have been steadily improving~\cite{li2018csrnet,wang2020distribution,ranasinghe2024crowddiff}. Current methods mainly use point-level positions or density maps as supervision, learning a nonlinear mapping from input images to the number of individuals~\cite{du2023redesigning}. While these methods have achieved impressive performance, their requirement for time-consuming and labor-intensive annotation limits the data available, and thus their robustness and generality.

To address this issue, some works have attempted to explore weakly supervised learning. For instance, the semi-supervised counting methods~\cite{wang2023self,chen2023multi} utilize a limited number of labeled samples to provide accurate supervision signals, and leverage unlabeled data to help understand diverse scenarios, thereby improving their accuracy and generalization ability. Nevertheless, these methods still rely on a number of annotated samples, and their performance largely depends on the quantity of these annotated samples. 
Unsupervised methods such as  \cite{babu2022completely, liang2023crowdclip} employ contrastive learning strategies to enable the network to discover the semantic relationships inherent in large volumes of unlabeled data. The performance of these unsupervised methods is generally lower than that of semi-supervised learning methods, however.

To avoid the heavy annotation burden typically associated with crowd counting, we propose an approach based on {\em rough labels} which provide only an approximate number of targets for each image. 
We term the task of estimating crowd numbers from images on the basis of rough labels as {\em Rough Crowd Counting}. It is inspired by humans' ability to make inexact quantity estimation and comparison from infancy, which may be related to the development of accurate numerical concepts and later mathematics in humans~\cite{cheyette2019primarily}. These rough labels help bypass the labor-intensive task of point annotations or precise count annotations.

We propose a novel solution to the problem posed above 
based on Progressive estimation for Rough Crowd Counting we label \method. 
The choice of CLIP is motivated by two key reasons: 1) CLIP is trained on a large-scale dataset of image-text pairs, which endows it with strong generalization capabilities in downstream tasks. This helps the model mitigate the impact of noise and labeling inconsistencies. 
2) Text descriptions provide rich supervisory signals for the model during training, allowing the model to better capture the nuances of visual concepts and their relationships. 
However, directly applying CLIP for crowd counting poses significant challenges. CLIP makes predictions by measuring the similarity between an image and a given text description. Since crowd counts can range from 0 to thousands, the original CLIP would need to perform thousands of comparisons before making a prediction, leading to severe inefficiency.

To address this problem, we propose a progressive estimation learning strategy, as shown in \cref{fig1}. Taking an image patch as an example, instead of matching text for all possible target counts, we predict each component digit of the count, starting with the highest place to the lowest place, \eg~hundreds place $\rightarrow$ tens place $\rightarrow$ units place. 
For example, when the number of targets is under 1,000, we categorize the hundreds place into 10 groups. Initially, the model predicts the hundreds digit; then, based on this prediction, it forecasts the tens digit, which also falls into one of 10 categories. Finally, with these two predicted digits, we proceed to estimate the units digit. In this way, we just need to perform 30 matches with the capability to estimate up to 999 objects, which significantly reduces the overhead of network inference. 
It is noteworthy that the number of matches of our method is linearly increased as the maximum of targets increases 10 times. 
Furthermore, we introduce a visual-language matching adapter to improve the matching accuracy. This adapter captures and retains the effective matching information of visual embedding and text embedding for refining visual outputs. Extensive experiments on  SHA~\cite{zhang2016single}, QNRF~\cite{idrees2018composition}, and JHU++~\cite{sindagi2020jhu}  demonstrate the favorable performance of the proposed method with comparison to state-of-the-art semi-supervised and unsupervised methods. In summary, our main contributions are as follows:
\begin{itemize}
	\item To the best of our knowledge, this is the first exploration of rough crowd counting. It avoids the time-consuming and labor-intensive point annotations and precise count annotations. This eases deployment in various scenarios.
	\item We propose a progressive estimation learning strategy, predicting the target count as digits from high to low place. This enables the utilization of board knowledge embedded in CLIP while significantly decreasing the computational cost during inference.
    \item We propose a visual-language matching adapter to optimize the key-value pairs by mining effective matching information of two modalities for improving the matching accuracy.
    \item Extensive experiments on three widely used benchmarks demonstrate the effectiveness of our method compared to several state-of-the-art weakly-/semi-superived methods.
\end{itemize}

\section{Related Work}
\subsection{Crowd Counting}
Crowd counting is a particularly challenging task due to factors such as severe occlusions, variations in scale, complex backgrounds, and uneven density distributions. To date, considerable efforts have been devoted to studying this task, resulting in significant progress. From the perspective of supervisory signals, crowd counting methods are typically categorized into three types: {\em fully-supervised}, {\em semi-supervised}, and {\em unsupervised}.

Fully-supervised counting methods use labeled data as training targets, with point annotation of objects being a common labeling strategy. Early methods often used Gaussian kernels to convolve point coordinates and generate density maps~\cite{zhang2016single}, which served as the supervision objective of the network. A key challenge with this approach is selecting an optimal Gaussian kernel size for generating the density map. To address this,~\cite{ma2019bayesian} introduced a Bayesian loss that calculates the expected count for each annotated point to guide the network. Another notable method, DM-Count~\cite{wang2020distribution}, computed the similarity between normalized prediction results and density maps using optimal transport. More recently, some methods shifted to point or box coordinate regression rather than probability distribution regression, as demonstrated in~\cite{sam2020locate,liang2022end,liu2023point}. While these fully-supervised methods have established a solid foundation for crowd counting research, they heavily rely on large amounts of annotated data, which is time-consuming and labor-intensive.

Semi-supervised and unsupervised counting methods provide effective strategies for reducing reliance on extensive annotated samples. Semi-supervised methods~\cite{wang2023self,chen2023multi,qian2024semi} work by assigning pseudo-labels to unlabeled images based on patterns learned from annotated images. This learning paradigm helps the network capture more robust features and improves generalization by leveraging both labeled and unlabeled data. However, these methods still depend on annotated data, and their performance is influenced by the amount of available labeled samples. Unsupervised methods~\cite{sam2019almost,babu2022completely}, on the other hand, do not use any annotated data. They learn to count objects based on predefined rules or optimization strategies. Although these methods can establish relationships between object counts, their performance typically lags behind that of semi-supervised learning methods. This indicates that entirely discarding human-annotated priors may impede effective model learning. To overcome this challenge, we propose using rough labels as learning targets. This approach circumvents the time-consuming and labor-intensive process of one-to-one annotation while utilizing existing prior information to enhance the model’s ability to accurately identify targets.

\subsection{Contrastive Language-image Pre-training}
Pre-training visual language models have been remarkably successful, providing a solid foundation for a wide range of downstream tasks~\cite{mo2023s}. Among these models, CLIP~\cite{radford2021learning} stands out as a prime example. It is pretrained on 400 million image-text pairs and learns a joint embedding between the two modalities. This model exhibits strong zero-shot transfer capabilities across various downstream tasks, and has attracted considerable attention.~\cite{kan2023knowledge} studied two complementary types of knowledge-aware prompts for the text encoder, aiming to reduce overfitting to seen classes and enhance generalization to unseen classes.

Recently, some works have started to explore the application of CLIP models to counting tasks, resulting in notable advancements. CrowdCLIP~\cite{liang2023crowdclip} utilized CLIP for unsupervised crowd counting and introduced a ranking loss that imposes ordinal information into the image-text similarity map.~\cite{paiss2023teaching} introduced a simple contrastive learning strategy that enables CLIP to perceive the number of targets, constructing negative samples by varying the number of targets in text prompts. This method builds upon the original contrastive learning strategy used in CLIP but restricts the counting range from one to ten, limiting its applicability to scenarios with larger numbers of targets. CLIP-Count~\cite{jiang2023clip} predicted density maps for open-vocabulary objects with text guidance in a zero-shot manner. Benefiting from the pre-trained knowledge of the CLIP model, it achieves good performance on multiple datasets, but this method still relies on a large amount of annotated data. Unlike these methods, we propose using rough labels as supervisory signals, avoiding the cumbersome process of dense point annotation, while also making it scalable to counting tasks involving different numbers of targets.

\section{Method}
\begin{figure*}[t]
	\centering
	\includegraphics[width=2.1\columnwidth]{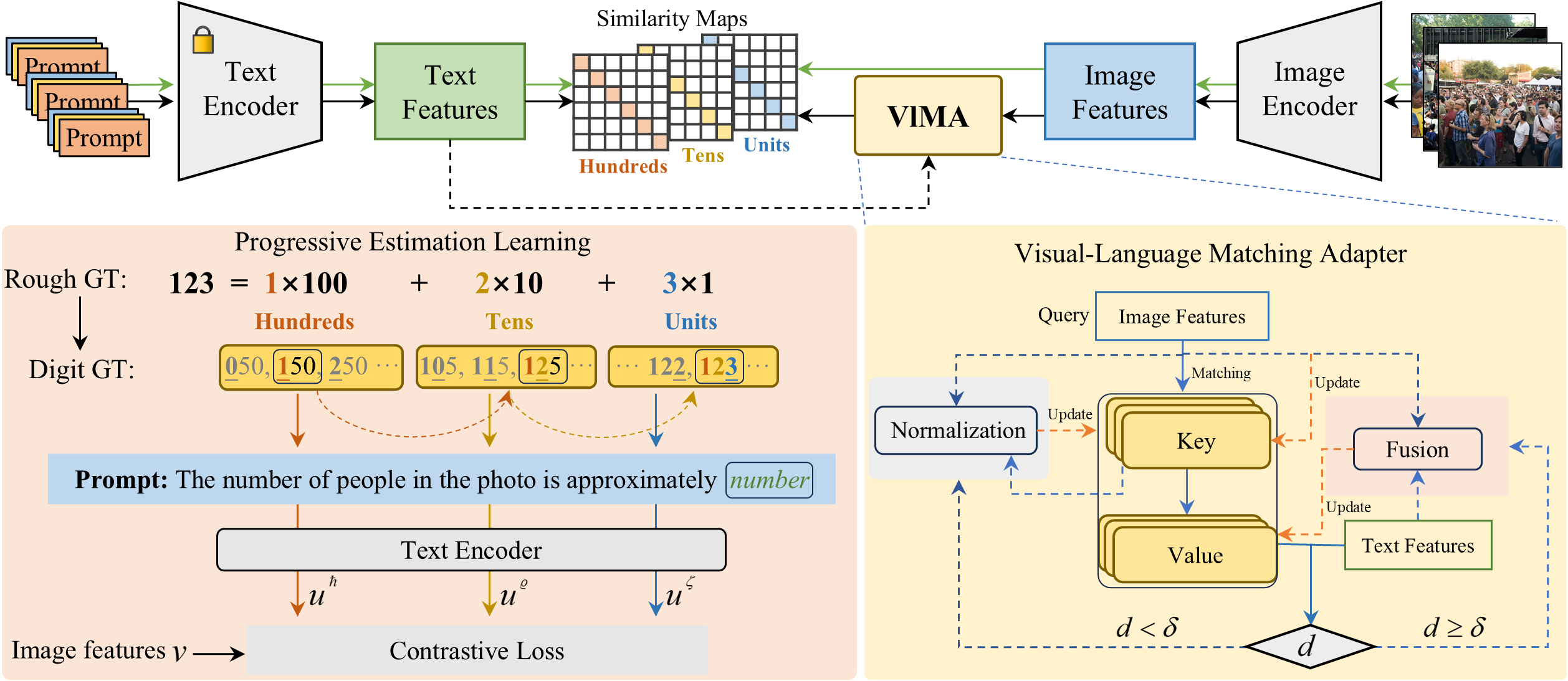}
	\caption{Overview of our proposed \method. We freeze the text encoder in pretrained CLIP and finetune the image encoder by a progressive estimation learning (PEL) strategy (Sec.~\ref{sec:PEL}). Then, a visual-language matching adapter (VlMA, Sec.~\ref{sec:VLMA}) is introduced to improve the image features for more reliable visual-to-text matching. The green straight arrows and the black straight arrows denote the stages of training PEL and VlMA sequentially.}
	\label{fig2}
	\vspace{-4mm}
\end{figure*}

In this paper, we introduce \method, a novel progressive CLIP framework for rough crowd counting. It leverages approximate quantity-estimated labels, referred to as rough labels,  for network supervision. We start with a brief overview of contrastive learning in CLIP and then explore the two key components of \method: {\em progressive estimation learning} and {\em visual-language matching adapter}. The overall architecture of our network is illustrated in \cref{fig2}.

\vspace{2mm}
\noindent \textbf{Preliminary of Contrastive Learning in CLIP}

Thanks to its robust generalization capabilities,  CLIP has demonstrated excellent performance on a variety of downstream tasks. CLIP uses contrastive learning to learn from image-text pairs to obtain a joint embedding space.
Given a batch of the image-text pairs $\{ {I_i},{T_i}\} _{i = 1}^N$, they are separately fed into the image encoder and the text encoder to obtain their embedding vector pairs $\{ {v_i},{u_i}\} _{i = 1}^N$. The network parameters are optimized by maximizing the similarity between positive image-text pairs (where an image and a text description share similar or the same semantics) and minimizing the similarity between negative pairs.Formarlly,  CLIP minimizes the following objective~\cite{zhang2022contrastive}:

\begin{equation}\label{eq1}
{\mathcal{L}_{clip}} = \frac{1}{{2N}}\sum\limits_{i = 1}^N {(\ell _i^{image}({v_i}} ,{u}) + \ell _i^{text}({u_i},{v})),
\end{equation}
where $\ell _i^{image}({v_i} ,{u})$ and $\ell _i^{text}({u_i},{v})$ denote the image-to-text and text-to-image contrastive losses, respectively:

\begin{equation}
\ell _i^{image}({v_i},{u}) =  - \log \frac{{\exp (\langle {v_i},{u_i}\rangle /\tau )}}{{\sum\limits_{n = 1}^N {\exp } (\langle {v_i},{u_n}\rangle /\tau )}},
\label{eq2}
\end{equation}

\begin{equation}
\ell _i^{text}({u_i},{v}) =  - \log \frac{{\exp (\langle {u_i},{v_i}\rangle /\tau )}}{{\sum\limits_{n = 1}^N {\exp } (\langle {u_i},{v_n}\rangle /\tau )}},
\label{eq3}
\end{equation}
where ${\langle {v_i},{u_i}\rangle }$ indicates the cosine similarity between $v_i$ and $u_i$, and $\tau$ denotes the temperature parameter.

During inference, for a given image, CLIP predicts its category
by comparing the similarity between the image and all category candidates in a format like ``This is a photo of xx'', and the one with the highest similarity is selected as prediction:  
\begin{equation}\label{eq4}
	P({v_i}) = {\arg \max} _{k \in \mathcal{M}}\left\langle {{v_i},u^k} \right\rangle,
\end{equation}
where $\mathcal{M}$ denotes the set of category names. 

\subsection{Progressive Estimation Learning}
\label{sec:PEL}

The approximate quantity estimation alleviates the heavy burden of annotating dense crowds, but also poses a significant challenge to robust learning due to discrepancies between estimated and actual counts. To address this issue, we utilize CLIP, known for its powerful generalization, as our learning framework. 
Our goal is to effectively explore the connection between visual and textual representations. 
A straightforward approach would be to train the model to associate visual inputs directly with text prompts that include estimated counts. For example, using prompts like  ``\texttt{The number of people in the photo is approximately $\it{number}$}'', where ``$\it{number}$'' corresponds to rough labels. 
While this method avoids additional computational load during training, it becomes inefficient during inference. 
This inefficiency arises because the model has to compare the query image against text prompts containing all possible target counts to determine the most accurate count, leading to a time-consuming inference process.

To tackle this issue, we propose a progressive estimation learning strategy. This method predicts target counts by estimating numbers from high to low digits, which simplifies the estimation task for the network and avoids the need to match with numerous text prompts. 
Specifically, we categorize rough labels into three-digit ranges—hundreds, tens, and units—and use these categories as supervision signals to train the network. 
For instance, if the rough groundtruth is 123, the label of the hundreds place would be set to 150, indicating that the actual count falls between 100 and 200. Using this information, the model then estimates the tens place, with the label set to 125, suggesting the count is between 120 and 130. Finally, the label of the units place directly uses the groundtruth as supervision. Despite real samples may contain more than 1,000 objects, the CLIP model processes only a fixed-size patch from the original image. Consequently, there is no need to set an excessively large counting range for the majority of counting scenarios. We have therefore set the counting range from 0 to 999. We utilize labels for the hundreds, tens, and units digits to generate text prompts, which are then fed into the text encoder to generate corresponding text features, i.e., $u_i^{\hbar}$, $u_i^{\varrho}$, and $u_i^{\zeta}$, and apply contrastive learning loss to optimize the network parameters:
\begin{equation}
{\mathcal{L}_{clip_{-}es}}{\text{ }} = \frac{1}{{2N}}\sum\limits_{k \in \mathcal{S}} {\sum\limits_{i = 1}^N ( } \ell _i^{image}({v_i},u^k) + \ell _i^{text}(u_i^k,{v})),
\end{equation}
where $\mathcal{S} = \{\hbar, \varrho, \zeta \}$. 

The advantage of this learning strategy is that during inference, we do not need to match against all possible object numbers. Instead, the model makes predictions by matching with a few of prompts generated from different labels within each digit range. 
Specifically, as shown in \cref{fig2}, during inference, we first estimate the hundreds digit by matching the image input with text prompts such as ``50, 150, ..., 950'' to determine the value of the hundreds digit. If the matched result is 150, we can conclude that the hundreds digit is 1. Based on this, we then predict the tens digit by matching the image input with text prompts generated from ``105, 115, ..., 195'' to determine the value of the tens digit. If the result is 125, then the tens digit is 2. Using these two digits, we match the input with ``120, 121, ..., 129'' to obtain the final prediction. The inference pseudocode is shown in \cref{alg1}. By following our proposed digit-matching strategy, for each query sample, we need only 30 matches to obtain the final count, which is just 3\% of the number of matches required when inferring all numerical values. This greatly speeds up the inference speed of CLIP for counting task.
\begin{algorithm}[tb]
\caption{Inference of Progressive Estimation Learning.}
\label{alg1}
\textbf{Input}: Image input: $I$, Text template: $T$\\
\textbf{Model}: Image encoder: $f( \cdot )$, Text encoder: $g( \cdot )$\\
\textbf{Output}: Estimation result: $P$
\begin{algorithmic}[1]
\STATE $number_{\hbar } \leftarrow \{50, 150, 250, \cdots, 950 \}$
\STATE $number_{\varrho} \leftarrow \{5, 15, 25, \cdots, 95 \}$
\STATE $number_{\zeta} \leftarrow \{0, 1, 2, \cdots, 9 \}$
\STATE $v \leftarrow f(I)$
\FOR{each $k \in \{\hbar, \varrho, \zeta\}$}
\IF{$k = \hbar$}
    \STATE $L \leftarrow number_{\hbar}$
\ELSIF{$k = \varrho$}
    \STATE $L \leftarrow number_{\varrho} + 100 \cdot P^{\hbar}$
\ELSE
    \STATE $L \leftarrow number_{\zeta} + 100 \cdot P^{\hbar} + 10 \cdot P^{\varrho}$
\ENDIF
\STATE \text{Fill $L$ into the description of $T$}
\STATE $u \leftarrow g(T)$
\STATE $P^k \leftarrow {\arg \max}_{k \in \mathcal{M}} \left \langle {{v},u^k} \right \rangle$    ~\# \cref{eq4} 
\ENDFOR
\STATE $P \leftarrow 100 \cdot P^{\hbar} + 10 \cdot P^{\varrho} + P^{\zeta}$
\STATE \textbf{return} $P$
\end{algorithmic}
\end{algorithm}

\subsection{Visual-Language Matching Adapter}
\label{sec:VLMA}

The image-text matching mechanism in CLIP plays a critical role in determining the final number of targets. To enhance the quality of two-modal matching, we further propose a visual-language matching adapter that captures the effective matching information from image and text embeddings. In other words, it memorizes effective representations by analyzing the relationship between image and text features during training, and then refines the visual features during inference to achieve a more accurate alignment with the corresponding textual features. The adapter consists of $M$ key-value pairs, denoted as \{$({key_{i}, value_{i}}), {i=1,2,...,M}$\}. The optimization of the adapter consists of two steps: a query step and an update step.  During the query phase, each visual encoding vector is measured against the keys to find the most similar value, which can be formulated as:

\begin{equation}
	value({v_i}) = value\left[ {\arg {{\max }_{j \in [0,M - 1]}}\left\langle {{v_i},key_j^T} \right\rangle } \right].
\end{equation}

 \begin{figure*}[!tb]
\centering
\includegraphics[width=1.8\columnwidth]{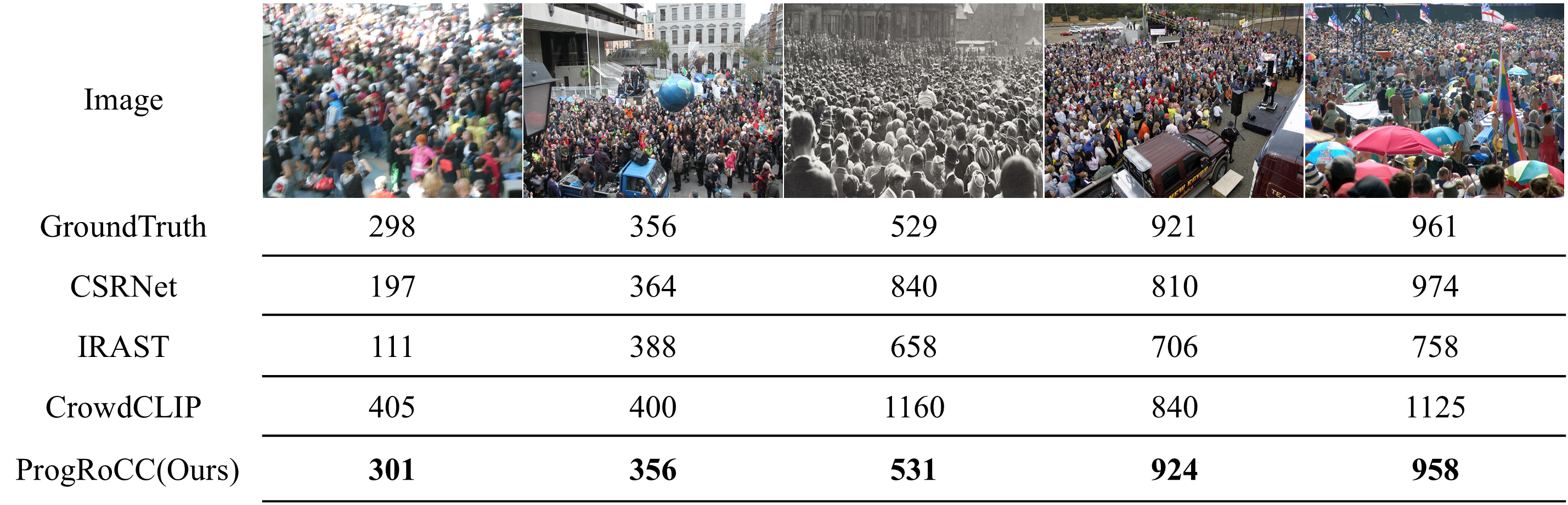}
\caption{Examples of predicted results on SHA.}
\label{fig3}
\end{figure*}

Further, we calculate the Euclidean distance $d_i$ between the value obtained from the query ($value({v_i})$) and the text encoding vector ($u_i$) to assess their relationship. Then, we can update the key-value pair in the adapter, allowing it to better characterize visually similar inputs. During the update process, we evaluate the relationship between the two using a threshold $\delta$, which guides us in updating the key-value pairs. Specifically, if the distance $d_i$ is less than  $\delta$, it indicates that the obtained value can accurately represent the features of the current visual input. In this case, we only need to update the key $key({v_i})$ to better align with the representative value. Conversely, if the distance is greater than or equal to $\delta$, it suggests that the value has weaker relevance to the visual input. Here we select the key-value pair that has remained unchanged for the longest duration for updating, to stimulate the representational potential of the adapter. Formally, if ${d_i} < \delta$, we update the selected key for the visual input $v_i$ as follows:
\begin{equation}
    key({v_i}) = \varphi \left( {key({v_i}) + {v_i}} \right),
\end{equation}
and if ${d_i} \geqslant \delta$, we perform the following update:
\begin{equation}
\left\{ {\begin{array}{*{20}{l}}
  {key(m) = {v_i},} \\ 
  {value(m) = {\lambda}{v_i} + (1-{\lambda}){u_i},} 
\end{array}\,\;} \right.
\end{equation}
where $\varphi(\cdot)$ indicates the normalization operation,  $\lambda $ is a scaling factor and $m$ denotes the index of the adapter to be updated. During inference, we use the value obtained from the query and perform a weighted average with the visual encoding vector. 

\begin{equation}
	{{\hat v}_i} = \frac{1}{2}\left( {{v_i} + value({v_i})} \right).
\end{equation}

This enhanced result ${{\hat v}_i}$ is then matched with the text encoding vector to obtain the final prediction.

\section{Experiment}
\subsection{Datasets and Evaluation Metrics}
We evaluate our method using three publicly crowd counting datasets: SHA~\cite{zhang2016single}, QNRF~\cite{idrees2018composition} and JHU++~\cite{sindagi2020jhu}.

\begin{itemize}
\item SHA~\cite{zhang2016single} consists of images randomly crawled from the Internet, comprising 482 images in total, with 300 images for training and 182 for testing.
\item QNRF~\cite{idrees2018composition} collects crowd images from diverse scenes, encompassing 1,535 images, of which 1,201 are used for training and 334 for testing.
\item JHU++~\cite{sindagi2020jhu} features images captured under various weather conditions, posing significant challenges for many methods. It includes a total of 4,372 images, with 2,722 images for training, 500 for validation, and the remaining 1,600 for testing.
\end{itemize}

Here, we utilize two widely-used evaluation metrics in crowd counting: Mean Absolute Error (MAE) and Mean Squared Error (MSE), to assess the model performance.

\begin{table*}[!tb]
\centering
\resizebox{0.8\linewidth}{!}{
\begin{tabular}{l|c|c|cc|cc|cc}
    \toprule
    \multicolumn{1}{l}{\multirow{2}{*}{Method}} & \multicolumn{1}{c}{\multirow{2}{*}{Label}} & \multicolumn{1}{c}{\multirow{2}{*}{Supervision}} & \multicolumn{2}{c}{SHA} & \multicolumn{2}{c}{QNRF} & \multicolumn{2}{c}{JHU++} \\
    \cmidrule{4-9}
    \multicolumn{1}{l}{} & \multicolumn{1}{c}{} & \multicolumn{1}{c}{} & MAE & \multicolumn{1}{c}{MSE} & MAE & \multicolumn{1}{c}{MSE} & MAE & \multicolumn{1}{c}{MSE}\\
    \midrule
    MCNN~\cite{zhang2016single} & \multirow{7}{*}{Point} & \multirow{7}{*}{Fully-supervised} & 110.2 & 173.2 & 277.0 & 426.0 & 188.9 & 483.4\\ 
    CSRNet~\cite{li2018csrnet} & & & 68.2 & 115.0 & - & - & 85.9 & 309.2\\
    DUBNet~\cite{oh2020crowd} & & & 64.6 & 106.8 & 105.6 & 180.5 & - & -\\
    LSC-CNN~\cite{sam2020locate} & & & 66.4 & 117.0 & 120.5 & 218.2 & - & -\\
    CSRNet+LA-Batch~\cite{zhou2021locality} & & & 65.8 & 103.6 & 113.0 & 210.0 & - & -\\
    AWCC-Net~\cite{huang2023counting} & & & 56.2 & 91.3 & 76.4 & 130.5 & 52.3 & 207.2\\
    HMoDE+REL~\cite{du2023redesigning} & & & 54.4 & 87.4 & - & - & 55.7 & 214.6\\
    \midrule
    MATT~\cite{lei2021towards} & \multirow{3}{*}{Count} & \multirow{3}{*}{Fully-supervised} & 80.1 & 129.4 & - & - & - & -\\
    TransCrowd~\cite{liang2022transcrowd} & & & 66.1 & 105.1 & 97.2 & 168.5 & 74.9 & 295.6\\
    DSFormer~\cite{hu2023densitytoken} & & & 64.0 & 100.5 & 94.2 & 167.9 & - & -\\
    \midrule
    \midrule
    GWTA~\cite{sam2019almost} & \multirow{3}{*}{Point} & \multirow{3}{*}{Semi-supervised} & 154.7 & 229.4 & - & - & - & -\\
    IRAST~\cite{liu2020semi} & & & 86.9 & 148.9 & 135.6 & 233.4 & - & -\\
    SSL-FT~\cite{wang2023self} & & & 82.1 & 132.9 & 151.0 & 259.1 & - & -\\
    \midrule
    CSS-CCNN~\cite{babu2022completely} & \multirow{3}{*}{None} & \multirow{3}{*}{Unsupervised} & 197.3 & 295.9 & 437.0 & 722.3 & 217.6 & 651.3\\
    CrowdCLIP~\cite{liang2023crowdclip} & & & 146.1 & 236.3 & 283.3 & 488.7 & 213.7 & 576.1\\
    SEEM~\cite{wan2024robust} & & & 102.6 & 176.3 & 182.3 & 289.9 & 102.7 & 360.7\\
    \midrule
    \method~(Ours) & Rough Label & Weakly-supervised & 70.0 & 112.6 & 114.8 & 204.7 & 92.5 & 365.4\\
    \bottomrule
\end{tabular}}
\caption{Comparison with state-of-the-art methods on the SHA, QNRF and JHU++ datasets.}
\label{tab1}
\vspace{-4mm}
\end{table*}

\subsection{Implementation Details}

All experiments are conducted with PyTorch framework. We utilize the ViT-B/16 as the image decoder, which has a fixed spatial resolution of 224$\times$224.  The batch size is set to 128. During the fine-tuning stage, we use the Adam optimizer with a learning rate of 1e-7. For the visual-language matching adapter, the number of key-value pairs $M$ is set to 3,000 by default. The hyper-parameters $\delta$ and $\lambda$ are set to 0.14 and 0.1, respectively. In all experiments, we use the same hyperparameters unless otherwise specified.  

For the generation of rough labels, we define an error sampling range around the true labels, from which we randomly sample $n$ labels to simulate estimates by $n$ different experts, where $n$ has a default value of 10. The error sampling range is used to simulate the uncertainty different annotators have when estimating the same sample, which is given as a percentage. For example, ±5\% of the true label means that rough labels are randomly selected within a range of ±5\% above or below the real number of people. During training, we sample a number between the maximum and minimum of these labels as the training label, which allows the network to perceive the uncertainty in the labels. For testing, we use the true labels to ensure a fair comparison with other methods.

\subsection{Comparison with the State-of-the-Art Methods}
\cref{tab1} reports the comparison results of our method with state-of-the-art methods. Therein, unsupervised and semi-supervised methods are the primary objects of comparison, as our method relies on a weakly supervised signal, which is an empirical approximation estimate for each training sample. \method~consistently and significantly surpasses all unsupervised methods by a large margin. Notably, compared to the best-performing method, SEEM~\cite{wan2024robust}, our method improves MAE by 31.8\% (and MSE by 36.1\%) on SHA, and MAE by 37.0\% (with a 29.4\% improvement in MSE) on QNRF. This underscores that the easily available priors, i.e., rough labels, can significantly enhance the network's ability to accurately estimate the number of objects in visual inputs. 

Compared to semi-supervised methods, we also achieve superior performance. Specifically, our method outperforms the recent method SSL-FT~\cite{wang2023self} by 14.7\% MAE (15.3 \% MSE) on SHA and 24.0\% MAE (21.0\% MSE) on QNRF. These methods rely on partial point annotations to help the network have a good initialization, which still depends heavily on manual annotations. In comparison, our method only requires estimating a number for the targets in each training sample, which is simpler and more effective. The results suggest that samples with rough labels can help the model better perceive the number of targets. Furthermore, although our method lags behind recent fully-supervised methods, it interestingly delivers comparable or even superior results on certain datasets. For instance,  our method is 1.8 MAE behind CSRNet~\cite{li2018csrnet} but outperforms it by 2.4 MSE on SHA. our method achieves results nearly identical to the recent method CSRNet+LA-Batch~\cite{zhou2021locality} on QNRF. Note that compared to pixel-level and count-level annotations, rough labels are easier to obtain, significantly reducing the cost of large-scale data annotation and offering greater potential for robust crowd counting. We further provide some qualitative visualizations to analyze the effectiveness of our method, as shown in \cref{fig3}. \method~performs well in scenes with varying densities.

\begin{table}[ht!]
\centering
\resizebox{\linewidth}{!}{
\begin{tabular}{ccc|c|cc|cc}
	\toprule
	\multicolumn{1}{c}{\multirow{2}{*}{Baseline}} & \multicolumn{1}{c}{\multirow{2}{*}{PEL}} & \multicolumn{1}{c}{\multirow{2}{*}{VlMA}} & \multicolumn{1}{c}{\multirow{2}{*}{FPS}} & \multicolumn{2}{c}{SHA} & \multicolumn{2}{c}{QNRF} \\
	\cmidrule{5-8}
	& & \multicolumn{1}{c}{} & \multicolumn{1}{c}{} & MAE & \multicolumn{1}{c}{MSE} & MAE & \multicolumn{1}{c}{MSE} \\
	\midrule
	$\checkmark$ & & & 5 &94.4  &150.6  & 121.9 & 213.2 \\
	\midrule
    & $\checkmark$ & & 56 & 72.1 & 117.1 & 117.0 & 205.5 \\
    & $\checkmark$ & $\checkmark$ & 55 & 70.0 & 112.6 & 114.8 & 204.7 \\
    \bottomrule
\end{tabular}}
\caption{Ablation study on model components.  PEL and VlMA denote progressive estimation learning and visual-language matching adapter, respectively.}
\label{tab20}
\end{table}

\begin{table}[ht!]
\centering
\resizebox{0.75\linewidth}{!}{
\begin{tabular}{c|cc}
	\toprule
	\multicolumn{1}{c}{Error Sampling Range} & MAE & MSE\\
	\midrule
	$({\text{GT - 5\% ,}}\,{\text{GT + 5\% }})$   & 117.0 & 205.5 \\
	$({\text{GT - 10\% ,}}\,{\text{GT + 10\% }})$ & 133.6 & 233.1 \\
	$({\text{GT - 15\% ,}}\,{\text{GT + 15\% }})$ & 121.9 & 228.7 \\
	$({\text{GT - 20\% ,}}\,{\text{GT + 20\% }})$ & 119.8 & 205.4 \\
	$({\text{GT - 30\% ,}}\,{\text{GT + 30\% }})$ & 124.3 & 223.9  \\
        $({\text{GT - 40\% ,}}\,{\text{GT + 40\% }})$ & 128.4 & 229.3  \\
        $({\text{GT - 50\% ,}}\,{\text{GT + 50\% }})$ & 124.9 & 224.0  \\
    \bottomrule
\end{tabular}
}
\caption{Ablation study on rough labels.}
\label{tab30}
\vspace{-2mm}
\end{table}

\begin{table}[ht!]
\centering
\resizebox{\linewidth}{!}{
\begin{tabular}{c|c|c|cc|cc}
	\toprule
	\multicolumn{1}{c}{\multirow{2}{*}{Method}} & \multicolumn{1}{c}{\multirow{2}{*}{Label}} & \multicolumn{1}{c}{\multirow{2}{*}{FPS}} & \multicolumn{2}{c}{SHA} & \multicolumn{2}{c}{QNRF} \\
	\cmidrule{4-7}
	\multicolumn{1}{c}{} & \multicolumn{1}{c}{} & \multicolumn{1}{c}{} & MAE & \multicolumn{1}{c}{MSE} & MAE & \multicolumn{1}{c}{MSE} \\
	\midrule
    \multirow{2}{*}{TransCrowd~\cite{liang2022transcrowd}} & Count & \multirow{2}{*}{167} & 66.1 & 105.1 & 97.2 & 168.5 \\
    & Rough Label & & 110.9 & 193.4 & 289.4 & 596.1 \\
	\midrule
    \multirow{2}{*}{CCTrans~\cite{tian2021cctrans}} & Count & \multirow{2}{*}{91} & 64.4 & 95.4 & 92.1 & 158.9 \\
    & Rough Label & & 84.3 & 118.8 & 122.1 & 188.4 \\
    \midrule
    \method~(Ours) & Rough Label & 55 & 70.0 & 112.6 & 114.8 & 204.7 \\
    \bottomrule
\end{tabular}}
\caption{Effects of rough labels on other methods.}
\label{tab30_1}
\end{table}

\subsection{Ablation Study}
In this part, we conduct a series of ablation studies to verify the effectiveness of our method on QNRF, unless otherwise specified.

\begin{table*}[!ht]
\centering
\renewcommand{\thetable}{5}
\resizebox{0.9\linewidth}{!}{
\begin{tabular}{l|cc|cc|cc|cc|cc|cc}
    \toprule
    \multicolumn{1}{l}{\multirow{2}{*}{Method}} & \multicolumn{2}{c}{$\left [0,100 \right )$} & \multicolumn{2}{c}{$\left [100,200 \right )$} & \multicolumn{2}{c}{$\left [200,300 \right )$} & \multicolumn{2}{c}{$\left [300,500 \right )$} & \multicolumn{2}{c}{$\left [500,800 \right )$} & \multicolumn{2}{c}{$\left [800,+\infty \right )$} \\
    \cmidrule{2-13}
    \multicolumn{1}{c}{} & MAE & \multicolumn{1}{c}{MSE} & MAE & \multicolumn{1}{c}{MSE} & MAE & \multicolumn{1}{c}{MSE} & MAE & \multicolumn{1}{c}{MSE} & MAE & \multicolumn{1}{c}{MSE} & MAE & \multicolumn{1}{c}{MSE} \\
    \midrule
    CCTrans~\cite{tian2021cctrans} & 21.8 & 22.4 & 27.6 & 39.8 & 49.4 & 60.7 & 80.9 & 118.5 & 121.4 & 141.7 & 228.9 & 299.2 \\
    \midrule
    TransCrowd~\cite{liang2022transcrowd} & 18.1 & 24.1 & 33.1 & 43.7 & 48.5 & 62.5 & 96.1 & 119.7 & 157.9 & 195.4 & 816.9 & 1107.7 \\
    \midrule
    \method~(Ours) & 10.6 & 13.3 & 25.1 & 41.6 & 32.7 & 46.0 & 65.5 & 96.8 & 121.0 & 161.1 & 263.2 & 355.1 \\
    \bottomrule
\end{tabular}}
\caption{Comparison of performance in different density intervals on QNRF.}
\label{tab_range}
\end{table*}

\begin{table*}[!ht]
\centering
\renewcommand{\thetable}{7}
\resizebox{0.9\linewidth}{!}{
\begin{tabular}{l|c|c|cc|cc|cc}
    \toprule
    \multicolumn{1}{l}{\multirow{2}{*}{Method}} & \multicolumn{1}{c}{\multirow{2}{*}{Label}} & \multicolumn{1}{c}{\multirow{2}{*}{Supervision}} & \multicolumn{2}{c}{SHA$\rightarrow$QNRF} & \multicolumn{2}{c}{SHA$\rightarrow$ JHU++} & \multicolumn{2}{c}{QNRF$\rightarrow$ SHA} \\
    \cmidrule{4-9}
    \multicolumn{1}{c}{} & \multicolumn{1}{c}{} & \multicolumn{1}{c}{} & MAE & \multicolumn{1}{c}{MSE} & MAE & \multicolumn{1}{c}{MSE} & MAE & \multicolumn{1}{c}{MSE} \\
    \midrule
    CSRNet~\cite{li2018csrnet} & \multirow{2}{*}{Point} & \multirow{2}{*}{Fully-supervised} & 216.0 & 402.6 & 144.7 & 600.1 & - & - \\
        HMoDE+REL~\cite{du2023redesigning} &  &  & 189.5 & 348.1 & 126.4 &530.9 & - & - \\
    \midrule
    IRAST~\cite{liu2020semi} & Point & Semi-supervised & 284.5 & 517.4 & 170.5 & 632.5 & 121.3 & 238.6 \\
    \midrule
    CSS-CCNN~\cite{babu2022completely} & \multirow{2}{*}{None} & \multirow{2}{*}{Unsupervised} & 472.4 & - & 251.3 & - & 235.7 & - \\
    CrowdCLIP~\cite{liang2023crowdclip} &  &  & 294.9 & 498.7 & 212.8 & 508.5 & 148.2 & 227.3 \\
    \midrule
    \method~(Ours) & Rough Label & Weakly-supervised & 171.0 & 290.0 & 116.3 & 379.2 & 93.8 & 147.0 \\
    \bottomrule
\end{tabular}}
\caption{Comparison on the transferability of our method and other methods under cross-dataset evaluation.}
\label{tab70}
\end{table*}


\subsubsection{Effectiveness of key components of \method.}
We first examine the key components of our method on SHA and QNRF. To show the benefits of our method, we establish a baseline that follows the original learning strategy of  CLIP, optimizing parameters using the contrastive loss defined in \cref{eq1}. During inference, the numerical range for text prompts is set to (0, 999), just as in our method. As shown in \cref{tab20}, with adding the progressive estimation learning strategy (PEL), the performance increases by 22.3 MAE and 38.0 MSE on SHA, and 4.9 MAE and 7.7 MSE on QNRF. Note that the inference speed increases by over 10 times at an input spatial resolution of $224 \times 224$ when using a 4090 GPU. We further apply visual-language matching adapter (VlMA) to PLE, the performance is improved continually on both datasets with adding almost no additional computational burden. This suggests that \method~is a more effective and efficient method compared to using the original CLIP for rough crowd counting.

\subsubsection{Effect of rough labels.}
For rough crowd counting, the expertise level of annotators is a significant factor affecting network performance. We quantitatively observe the impact of experts' annotation accuracy on the results by utilizing the range of sampling errors. As shown in \cref{tab30}, the best performance is achieved at an error sampling rate of ±5\%, with 117.0 MAE and 205.5 MSE. Performance declines as the error sampling range increases, due to the increased difficulty for the network to learn from large-span estimated labels. Additionally, performance remains relatively stable at error sampling rates between ±15\% and ±50\%, indicating that our progressive learning method effectively mitigates the impact of noisy estimated labels on network performance. Here we select the error sampling rate of ±5\% as our experimental setting.

To validate the superiority of our method in handling rough labels, we use them as supervision signals for training the recent methods, TransCrowd~\cite{tian2021cctrans} and CCTrans~\cite{tian2021cctrans}.  \cref{tab30_1} shows the results of the comparative experiment. From the table, it can be observed that compared to count-level labels, rough labels have a significant negative impact on the learning performance of the two methods. In contrast, our method demonstrates superior performance when dealing with such highly nosiy lables. Specifically, compared to the second-best method, CCTrans, our method achieves a 17\% improvement in MAE on SHA and an 18.6\% improvement on QNRF. We also notice that our method underperforms CCTrans in terms of MSE on QNRF. We speculate that this is because the dataset containing many samples with highly dense targets, while the predicted upper bound  of our network is set to less than 1000, restricting its performance in scenarios with a large number of targets. To further illustrate this, we evaluate the model’s performance across different density intervals in \cref{tab_range}. The results indicate that while our method almost performs well in the first four density intervals, its MSE performance declines in high-density scenarios. In summary, the results demonstrate that our method effectively addresses the challenge of learning from noisy labels.


\subsection{Cross-dataset Validation}
To explore the generalization performance of \method, we conduct the experiments on the transferability of our method and other methods under cross-dataset evaluation. The experimental results, as shown in \cref{tab70}, demonstrate the strong  transferability of our method. Notably, despite our model utilizes rough labels as supervision, our proposed method outperforms the SOTA fully-supervised methods, CSRNet~\cite{li2018csrnet} and HMoDE+REL~\cite{du2023redesigning}. Specially, compared to the best-performing fully-supervisied method HMoDE+REL, our method achieves a performance improvement of 9.8\% MAE (16.7\% MSE) and 8.0\% MAE (28.6\% MSE) when adapting the results trained on SHA to QNRF and JHU++, respectively, while our results being lower than that of HMoDE+REL on SHA. Compared with semi-supervised and unsupervised methods, our method significantly outperforms them on various transfer settings. These impressive results emphasize the good generalization of our method.

\section{Conclusion}
In this paper, we introduce a new counting paradigm called {\em rough crowd counting}. Rather than relying on time-consuming and labor-intensive annotations, we propose using approximate estimated counts as labels, which can provide valuable information while less precise than point annotations. To effectively leverage them, we propose a progressive CLIP framework, \method, to quickly adapt to the counting task. This framework features two key designs: a progressive estimation learning strategy and a visual-language matching adapter. Our method substantially boosts network performance and enhances inference efficiency compared to directly using the original CLIP model. Extensive experiments on crowd counting datasets demonstrate that our method achieves superior counting performance and exhibits strong generalization capabilities. We hope that our work offers a fresh perspective on the counting task.
{
    \small
    \bibliographystyle{ieeenat_fullname}
    \bibliography{main}

\begin{thebibliography}{33}
\providecommand{\natexlab}[1]{#1}
\providecommand{\url}[1]{\texttt{#1}}
\expandafter\ifx\csname urlstyle\endcsname\relax
  \providecommand{\doi}[1]{doi: #1}\else
  \providecommand{\doi}{doi: \begingroup \urlstyle{rm}\Url}\fi

\bibitem[Babu~Sam et~al.(2022)Babu~Sam, Agarwalla, Joseph, Sindagi, Babu, and
  Patel]{babu2022completely}
Deepak Babu~Sam, Abhinav Agarwalla, Jimmy Joseph, Vishwanath~A Sindagi,
  R~Venkatesh Babu, and Vishal~M Patel.
\newblock Completely self-supervised crowd counting via distribution matching.
\newblock In \emph{European Conference on Computer Vision}, pages 186--204.
  Springer, 2022.

\bibitem[Chen and Wang(2023)]{chen2023multi}
Jiwei Chen and Zengfu Wang.
\newblock Multi-task semi-supervised crowd counting via global to local
  self-correction.
\newblock \emph{Pattern Recognition}, 140:\penalty0 109506, 2023.

\bibitem[Cheyette and Piantadosi(2019)]{cheyette2019primarily}
Samuel~J Cheyette and Steven~T Piantadosi.
\newblock A primarily serial, foveal accumulator underlies approximate
  numerical estimation.
\newblock \emph{Proceedings of the National Academy of Sciences}, 116\penalty0
  (36):\penalty0 17729--17734, 2019.

\bibitem[Du et~al.(2023)Du, Shi, Deng, and Zafeiriou]{du2023redesigning}
Zhipeng Du, Miaojing Shi, Jiankang Deng, and Stefanos Zafeiriou.
\newblock Redesigning multi-scale neural network for crowd counting.
\newblock \emph{IEEE Transactions on Image Processing}, 32:\penalty0
  3664--3678, 2023.

\bibitem[Hu et~al.(2023)Hu, Wang, and Li]{hu2023densitytoken}
Zaiyi Hu, Binglu Wang, and Xuelong Li.
\newblock Densitytoken: Weakly-supervised crowd counting with density
  classification.
\newblock In \emph{IEEE International Conference on Acoustics, Speech and
  Signal Processing}, pages 1--5. IEEE, 2023.

\bibitem[Huang et~al.(2023)Huang, Chen, Chiang, Kuo, and
  Yang]{huang2023counting}
Zhi-Kai Huang, Wei-Ting Chen, Yuan-Chun Chiang, Sy-Yen Kuo, and Ming-Hsuan
  Yang.
\newblock Counting crowds in bad weather.
\newblock In \emph{Proceedings of the IEEE/CVF International Conference on
  Computer Vision}, pages 23308--23319, 2023.

\bibitem[Idrees et~al.(2018)Idrees, Tayyab, Athrey, Zhang, Al-Maadeed, Rajpoot,
  and Shah]{idrees2018composition}
Haroon Idrees, Muhmmad Tayyab, Kishan Athrey, Dong Zhang, Somaya Al-Maadeed,
  Nasir Rajpoot, and Mubarak Shah.
\newblock Composition loss for counting, density map estimation and
  localization in dense crowds.
\newblock In \emph{Proceedings of the European Conference on Computer Vision},
  pages 532--546, 2018.

\bibitem[Jiang et~al.(2023)Jiang, Liu, and Chen]{jiang2023clip}
Ruixiang Jiang, Lingbo Liu, and Changwen Chen.
\newblock Clip-count: Towards text-guided zero-shot object counting.
\newblock In \emph{ACM International Conference on Multimedia}, pages
  4535--4545, 2023.

\bibitem[Kan et~al.(2023)Kan, Wang, Lu, Zhen, Guan, and
  Zheng]{kan2023knowledge}
Baoshuo Kan, Teng Wang, Wenpeng Lu, Xiantong Zhen, Weili Guan, and Feng Zheng.
\newblock Knowledge-aware prompt tuning for generalizable vision-language
  models.
\newblock In \emph{Proceedings of the IEEE/CVF International Conference on
  Computer Vision}, pages 15670--15680, 2023.

\bibitem[Lei et~al.(2021)Lei, Liu, Zhang, and Liu]{lei2021towards}
Yinjie Lei, Yan Liu, Pingping Zhang, and Lingqiao Liu.
\newblock Towards using count-level weak supervision for crowd counting.
\newblock \emph{Pattern Recognition}, 109:\penalty0 107616, 2021.

\bibitem[Li et~al.(2018)Li, Zhang, and Chen]{li2018csrnet}
Yuhong Li, Xiaofan Zhang, and Deming Chen.
\newblock Csrnet: Dilated convolutional neural networks for understanding the
  highly congested scenes.
\newblock In \emph{Proceedings of the IEEE Conference on Computer Vision and
  Pattern Recognition}, pages 1091--1100, 2018.

\bibitem[Liang et~al.(2022{\natexlab{a}})Liang, Chen, Xu, Zhou, and
  Bai]{liang2022transcrowd}
Dingkang Liang, Xiwu Chen, Wei Xu, Yu Zhou, and Xiang Bai.
\newblock Transcrowd: Weakly-supervised crowd counting with transformers.
\newblock \emph{Science China Information Sciences}, 65\penalty0 (6):\penalty0
  160104, 2022{\natexlab{a}}.

\bibitem[Liang et~al.(2022{\natexlab{b}})Liang, Xu, and Bai]{liang2022end}
Dingkang Liang, Wei Xu, and Xiang Bai.
\newblock An end-to-end transformer model for crowd localization.
\newblock In \emph{European Conference on Computer Vision}, pages 38--54.
  Springer, 2022{\natexlab{b}}.

\bibitem[Liang et~al.(2023)Liang, Xie, Zou, Ye, Xu, and
  Bai]{liang2023crowdclip}
Dingkang Liang, Jiahao Xie, Zhikang Zou, Xiaoqing Ye, Wei Xu, and Xiang Bai.
\newblock Crowdclip: Unsupervised crowd counting via vision-language model.
\newblock In \emph{Proceedings of the IEEE/CVF Conference on Computer Vision
  and Pattern Recognition}, pages 2893--2903, 2023.

\bibitem[Liu et~al.(2023)Liu, Lu, Cao, and Liu]{liu2023point}
Chengxin Liu, Hao Lu, Zhiguo Cao, and Tongliang Liu.
\newblock Point-query quadtree for crowd counting, localization, and more.
\newblock In \emph{Proceedings of the IEEE/CVF International Conference on
  Computer Vision}, pages 1676--1685, 2023.

\bibitem[Liu et~al.(2020)Liu, Liu, Wang, Zhang, and Lei]{liu2020semi}
Yan Liu, Lingqiao Liu, Peng Wang, Pingping Zhang, and Yinjie Lei.
\newblock Semi-supervised crowd counting via self-training on surrogate tasks.
\newblock In \emph{European Conference on Computer Vision}, pages 242--259.
  Springer, 2020.

\bibitem[Ma et~al.(2019)Ma, Wei, Hong, and Gong]{ma2019bayesian}
Zhiheng Ma, Xing Wei, Xiaopeng Hong, and Yihong Gong.
\newblock Bayesian loss for crowd count estimation with point supervision.
\newblock In \emph{Proceedings of the IEEE/CVF International Conference on
  Computer Vision}, pages 6142--6151, 2019.

\bibitem[Mo et~al.(2023)Mo, Kim, Lee, and Shin]{mo2023s}
Sangwoo Mo, Minkyu Kim, Kyungmin Lee, and Jinwoo Shin.
\newblock S-clip: Semi-supervised vision-language learning using few specialist
  captions.
\newblock \emph{Advances in Neural Information Processing Systems},
  36:\penalty0 61187--61212, 2023.

\bibitem[Oh et~al.(2020)Oh, Olsen, and Ramamurthy]{oh2020crowd}
Min-hwan Oh, Peder Olsen, and Karthikeyan~Natesan Ramamurthy.
\newblock Crowd counting with decomposed uncertainty.
\newblock In \emph{Proceedings of the AAAI Conference on Artificial
  Intelligence}, pages 11799--11806, 2020.

\bibitem[Paiss et~al.(2023)Paiss, Ephrat, Tov, Zada, Mosseri, Irani, and
  Dekel]{paiss2023teaching}
Roni Paiss, Ariel Ephrat, Omer Tov, Shiran Zada, Inbar Mosseri, Michal Irani,
  and Tali Dekel.
\newblock Teaching clip to count to ten.
\newblock In \emph{Proceedings of the IEEE/CVF International Conference on
  Computer Vision}, pages 3170--3180, 2023.

\bibitem[Qian et~al.(2024)Qian, Hong, Guo, Arandjelovi{\'c}, and
  Donovan]{qian2024semi}
Yifei Qian, Xiaopeng Hong, Zhongliang Guo, Ognjen Arandjelovi{\'c}, and Carl~R
  Donovan.
\newblock Semi-supervised crowd counting with contextual modeling: facilitating
  holistic understanding of crowd scenes.
\newblock \emph{IEEE Transactions on Circuits and Systems for Video
  Technology}, 2024.

\bibitem[Radford et~al.(2021)Radford, Kim, Hallacy, Ramesh, Goh, Agarwal,
  Sastry, Askell, Mishkin, Clark, et~al.]{radford2021learning}
Alec Radford, Jong~Wook Kim, Chris Hallacy, Aditya Ramesh, Gabriel Goh,
  Sandhini Agarwal, Girish Sastry, Amanda Askell, Pamela Mishkin, Jack Clark,
  et~al.
\newblock Learning transferable visual models from natural language
  supervision.
\newblock In \emph{International Conference on Machine Learning}, pages
  8748--8763. PMLR, 2021.

\bibitem[Ranasinghe et~al.(2024)Ranasinghe, Nair, Bandara, and
  Patel]{ranasinghe2024crowddiff}
Yasiru Ranasinghe, Nithin~Gopalakrishnan Nair, Wele Gedara~Chaminda Bandara,
  and Vishal~M Patel.
\newblock Crowddiff: Multi-hypothesis crowd density estimation using diffusion
  models.
\newblock In \emph{Proceedings of the IEEE/CVF Conference on Computer Vision
  and Pattern Recognition}, pages 12809--12819, 2024.

\bibitem[Sam et~al.(2019)Sam, Sajjan, Maurya, and Babu]{sam2019almost}
Deepak~Babu Sam, Neeraj~N Sajjan, Himanshu Maurya, and R~Venkatesh Babu.
\newblock Almost unsupervised learning for dense crowd counting.
\newblock In \emph{Proceedings of the AAAI Conference on Artificial
  Intelligence}, pages 8868--8875, 2019.

\bibitem[Sam et~al.(2021)Sam, Peri, Sundararaman, Kamath, and
  Babu]{sam2020locate}
Deepak~Babu Sam, Skand~Vishwanath Peri, Mukuntha~Narayanan Sundararaman, Amogh
  Kamath, and R~Venkatesh Babu.
\newblock Locate, size, and count: accurately resolving people in dense crowds
  via detection.
\newblock \emph{IEEE Transactions on Pattern Analysis and Machine
  Intelligence}, 43\penalty0 (8):\penalty0 2739--2751, 2021.

\bibitem[Sindagi et~al.(2020)Sindagi, Yasarla, and Patel]{sindagi2020jhu}
Vishwanath~A Sindagi, Rajeev Yasarla, and Vishal~M Patel.
\newblock Jhu-crowd++: Large-scale crowd counting dataset and a benchmark
  method.
\newblock \emph{IEEE Transactions on Pattern Analysis and Machine
  Intelligence}, 44\penalty0 (5):\penalty0 2594--2609, 2020.

\bibitem[Tian et~al.(2021)Tian, Chu, and Wang]{tian2021cctrans}
Ye Tian, Xiangxiang Chu, and Hongpeng Wang.
\newblock Cctrans: Simplifying and improving crowd counting with transformer.
\newblock \emph{arXiv preprint arXiv:2109.14483}, 2021.

\bibitem[Wan et~al.(2024)Wan, Wu, Lin, and Chan]{wan2024robust}
Jia Wan, Qiangqiang Wu, Wei Lin, and Antoni~B Chan.
\newblock Robust unsupervised crowd counting and localization with adaptive
  resolution sam.
\newblock \emph{arXiv preprint arXiv:2402.17514}, 2024.

\bibitem[Wang et~al.(2020)Wang, Liu, Samaras, and Nguyen]{wang2020distribution}
Boyu Wang, Huidong Liu, Dimitris Samaras, and Minh~Hoai Nguyen.
\newblock Distribution matching for crowd counting.
\newblock \emph{Advances in Neural Information Processing Systems},
  33:\penalty0 1595--1607, 2020.

\bibitem[Wang et~al.(2023)Wang, Hao, Hu, Chen, Chen, and Wu]{wang2023self}
Rui Wang, Yixue Hao, Long Hu, Jincai Chen, Min Chen, and Di Wu.
\newblock Self-supervised learning with data-efficient supervised fine-tuning
  for crowd counting.
\newblock \emph{IEEE Transactions on Multimedia}, 25:\penalty0 1538--1546,
  2023.

\bibitem[Zhang et~al.(2016)Zhang, Zhou, Chen, Gao, and Ma]{zhang2016single}
Yingying Zhang, Desen Zhou, Siqin Chen, Shenghua Gao, and Yi Ma.
\newblock Single-image crowd counting via multi-column convolutional neural
  network.
\newblock In \emph{Proceedings of the IEEE Conference on Computer Vision and
  Pattern Recognition}, pages 589--597, 2016.

\bibitem[Zhang et~al.(2022)Zhang, Jiang, Miura, Manning, and
  Langlotz]{zhang2022contrastive}
Yuhao Zhang, Hang Jiang, Yasuhide Miura, Christopher~D Manning, and Curtis~P
  Langlotz.
\newblock Contrastive learning of medical visual representations from paired
  images and text.
\newblock In \emph{Machine Learning for Healthcare Conference}, pages 2--25.
  PMLR, 2022.

\bibitem[Zhou et~al.(2022)Zhou, Zhang, Du, Peng, Fang, Xiao, and
  Zhu]{zhou2021locality}
Joey~Tianyi Zhou, Le Zhang, Jiawei Du, Xi Peng, Zhiwen Fang, Zhe Xiao, and
  Hongyuan Zhu.
\newblock Locality-aware crowd counting.
\newblock \emph{IEEE Transactions on Pattern Analysis and Machine
  Intelligence}, 44\penalty0 (7):\penalty0 3602--3613, 2022.

\end{thebibliography}
}


\end{document}